# Fault Diagnosis of 3D-Printed Scaled Wind Turbine Blades [‹]

Luis Miguel Esquivel-Sancho[1r0000´0002´2591´8219s], Maryam Ghandchi Tehrani[1r0000´0002´0824´4937s], Mauricio Muñoz-Arias[1r0000´0003´0338´8285s], and Mahmoud Askari[1r0000´0001´5135´0850s]

Faculty of Science and Engineering, University of Groningen, Groningen, The Netherlands. l.m.esquivel.sancho@rug.nl

**Abstract.** This study presents an integrated methodology for fault detection in wind turbine blades using 3D-printed scaled models, finite element simulations, experimental modal analysis, and machine learning techniques. A scaled model of the NREL 5MW blade was fabricated using 3D printing, and crack-type damages were introduced at critical locations. Finite Element Analysis was employed to predict the impact of these damages on the natural frequencies, with the results validated through controlled hammer impact tests. Vibration data was processed to extract both time-domain and frequency-domain features, and key discriminative variables were identified using statistical analyses (ANOVA). Machine learning classifiers, including Support Vector Machine and K-Nearest Neighbors, achieved classification accuracies exceeding 94%. The results revealed that vibration modes 3, 4, and 6 are particularly sensitive to structural anomalies for this blade. This integrated approach confirms the feasibility of combining numerical simulations with experimental validations and paves the way for structural health monitoring systems in wind energy applications.

## 1 INTRODUCTION

The structural analysis of wind turbine blades is crucial for ensuring their reliability, especially as designs grow in size and complexity. These blades face harsh environmental conditions and are made from advanced composite materials, making fault detection challenging [7], [10]. Structural analysis is essential for identifying damage, such as cracks or degradation, which can impact performance. Effective structural analysis helps prevent failure and reduces maintenance costs by identifying weaknesses early [3].

Wind turbine blades are exposed to harsh environmental conditions and mechanical stresses, making them susceptible to faults that affect their performance and structural integrity. Common fault locations include the leading edge,

‹ Supported by University of Groningen

trailing edge, and root or mid-span sections. Cracks are frequently observed in the mid-span and root regions. Mid-span cracks result from cyclic loading and fatigue-induced bending, weakening the blade over time [10], [6]. Root cracks develop near the hub attachment, where stress concentrations are highest, often exacerbated by manufacturing defects such as improper bonding [10], [5].

Monitoring these structural damages is crucial for ensuring blade reliability. Vibration-based fault detection is a key approach in this context, as vibration signals capture the dynamic behavior of the blades and can reveal imbalances or hidden damage. Vibration analysis identifies damage-sensitive features, such as resonant frequencies and mode shapes, through techniques such as modal analysis. Modal analysis determines natural frequencies, mode shapes, and damping ratios, key characteristics for detecting structural changes that could indicate damage or deterioration [10], [5]. The adoption of machine learning in modal analysis offers precise and efficient detection. Techniques such as support vector machines can analyze complex data, achieving high accuracy rates [13], [1]. Despite their potential, real-world applications face challenges related to high computational demands and the need for large datasets [13].

This study presents an integrated approach for damage detection in wind turbine blades by combining numerical simulations, experimental modal analysis, and machine learning. Utilizing a scaled, 3D-printed NREL 5MW blade model [2] selected for its ability to produce highly repeatable samples, various damages are introduced, and their impacts on modal frequencies are analyzed. Time-domain and frequency-domain features are extracted and assessed for their discriminative power, focusing on the shifts in resonance frequencies of key vibration modes. The novelty of this work lies in the deep analysis of the blade's dynamic behavior across multiple damage configurations, employing vibration-based analysis to evaluate different types of damage and classify them.

This paper is structured as follows: Section 2 details the methodology, including the simulations, experiments, and data processing techniques used to analyze the turbine blade behavior. Furthermore, Section 3 presents the results and discusses their implications for fault detection and system performance. Finally, Section 4 concludes the paper, summarizing the findings and proposing future research directions.

## 2 Methodology

### 2.1 Physical Model and Damages

The study is based on the scaled-down NREL 5MW wind turbine blade, which is a widely recognized benchmark in the wind energy industry and academic research. It is divided into 17 sections, each with specific airfoil profiles. These sections include cylindrical and DU (Delft University) series airfoils, as well as NACA (National Advisory Committee for Aeronautics) profiles towards the tip of the blade [2]. The blade was modelled in CAD software and scaled to a length of 300 mm, the maximum size that could be 3D-printed. The blade was fabricated using a BambuLab 3D printer equipped with a 0.4 mm nozzle, employing Matte White PLA filament; PLA was chosen for its ease of manufacturing and mechanical properties. Printing was conducted vertically relative to the airfoil's chord line, using a 100% infill density, a layer height of 0.08 mm, and high-quality settings. A healthy blade was created as the baseline, alongside five distinct crack-type damages applied to critical regions identified in the literature [8], [13], such as the root, transition, and middle zones (Table 1, Figure 1). For each damage type, five blade units were printed under consistent conditions.

Table 1: Summary of blade damage types and their geometric parameters (Position, Length, Thickness, and Depth).

| Label | Type | Position (mm) | Length (mm) | Thickness (mm) | Depth (mm) |
|---|---|---|---|---|---|
| Healthy | No Damages | – | – | – | – |
| Damage 1 | Transverse crack | 15 | 15 | 2 | 5 |
| Damage 2 | Trailing edge transverse crack | 70 | 8 | 1 | Through |
| Damage 3 | Transverse crack | 70 | 16 | 2 | 1 |
| Damage 4 | Transverse crack | 50 | 10 | 2 | 2.5 |
| Damage 5 | Leading edge longitudinal crack | 65 | 20 | 2 | 8 |

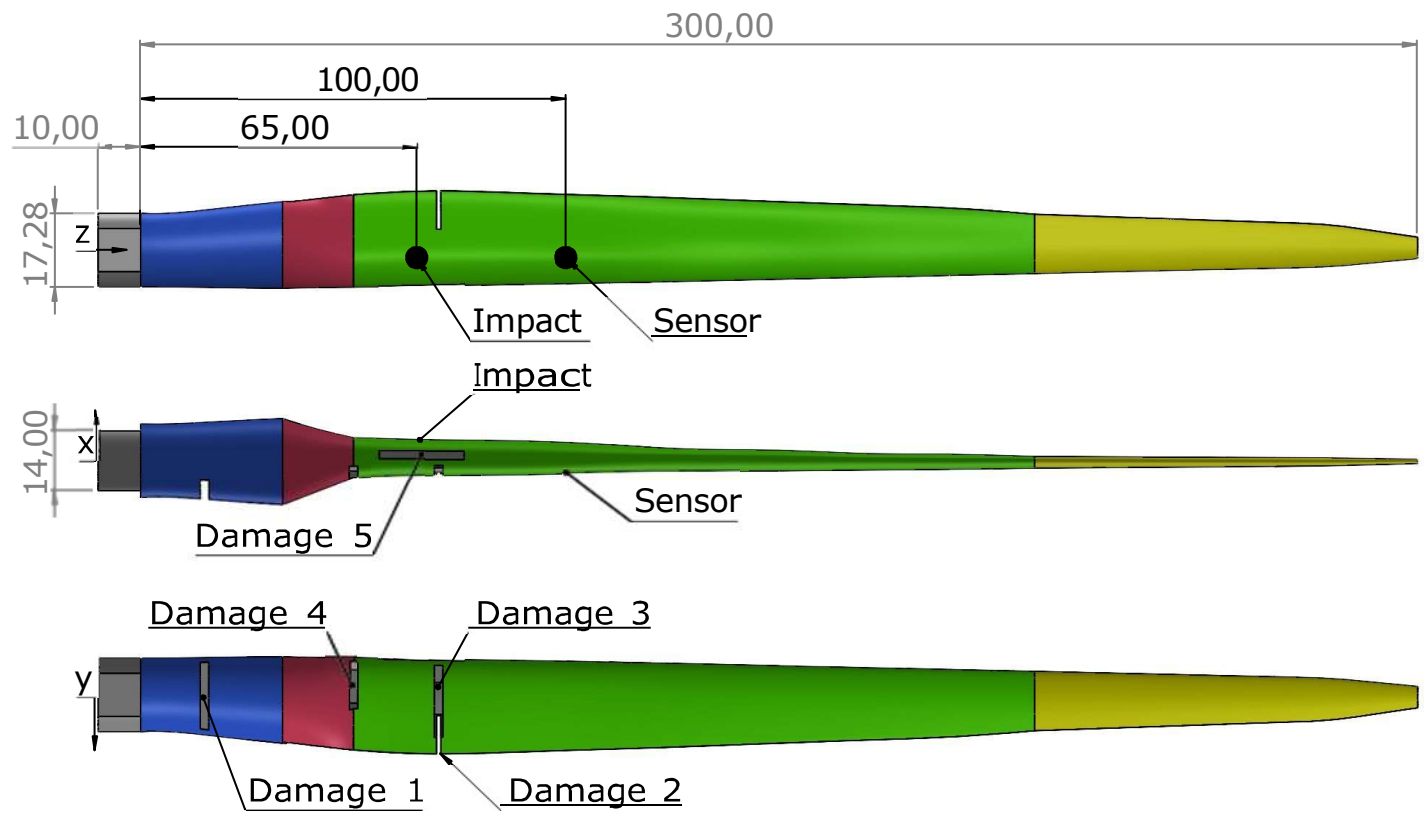

Fig. 1: Blade geometry of the scaled NREL 5MW blade, showing damage locations, impact and sensor positions, and XYZ coordinates. The gray area marks the fixation zone, blue the root, red the transition, green the midsection, and yellow the tip.

## 2.2 Numerical Simulation

Finite Element Analysis was conducted in MATLAB to predict the resonance frequencies and vibration modes of the blade for healthy and faulted specimens before doing the test. In the simulation study, a hammer test with a 40 N impact force over a time span of 0.1 ms to 0.2 ms was applied 65 mm from the blade root. Acceleration responses were recorded at a sensor located 100 mm from the root.

The finite element mesh was generated with maximum and minimum element sizes of 0.008 m and 0.003 m, respectively, with quadratic geometry, for a total of 3ˆ11485 nodes and 10ˆ6256 elements for the healthy blade. The blade material was modeled using the following properties: a density of 1124.6 kg/m$^3$, damping ratio of 0.015, Young's modulus of 2.55 GPa, and a Poisson's ratio of 0.35. The material properties data were based on [11], [9], then slightly adjusted to be consistent with the reality of the structure due to manufacturing characteristics. The simulation time was set to 0.5 seconds with 5000 steps for transient analysis.

### 2.3 Experimental Setup

The experimental setup involved securing the blades horizontally in a bench vise on an anti-vibration table (Figure 2). A 4507 B 004 accelerometer and a Dynapulse impulse hammer measured acceleration and impact force, respectively, with signals processed by a PCB Piezotronics 482C unit and recorded via a dSpace MicroLabBox. Each test consisted of a single 40 ˘ 10 N impact, with data collected for 5.5 s (0.5 s pre- and 5 s post-impact). Five blades per fault condition were tested 20 times, and results were saved in `.mat` format, organized by damage type and unit number.

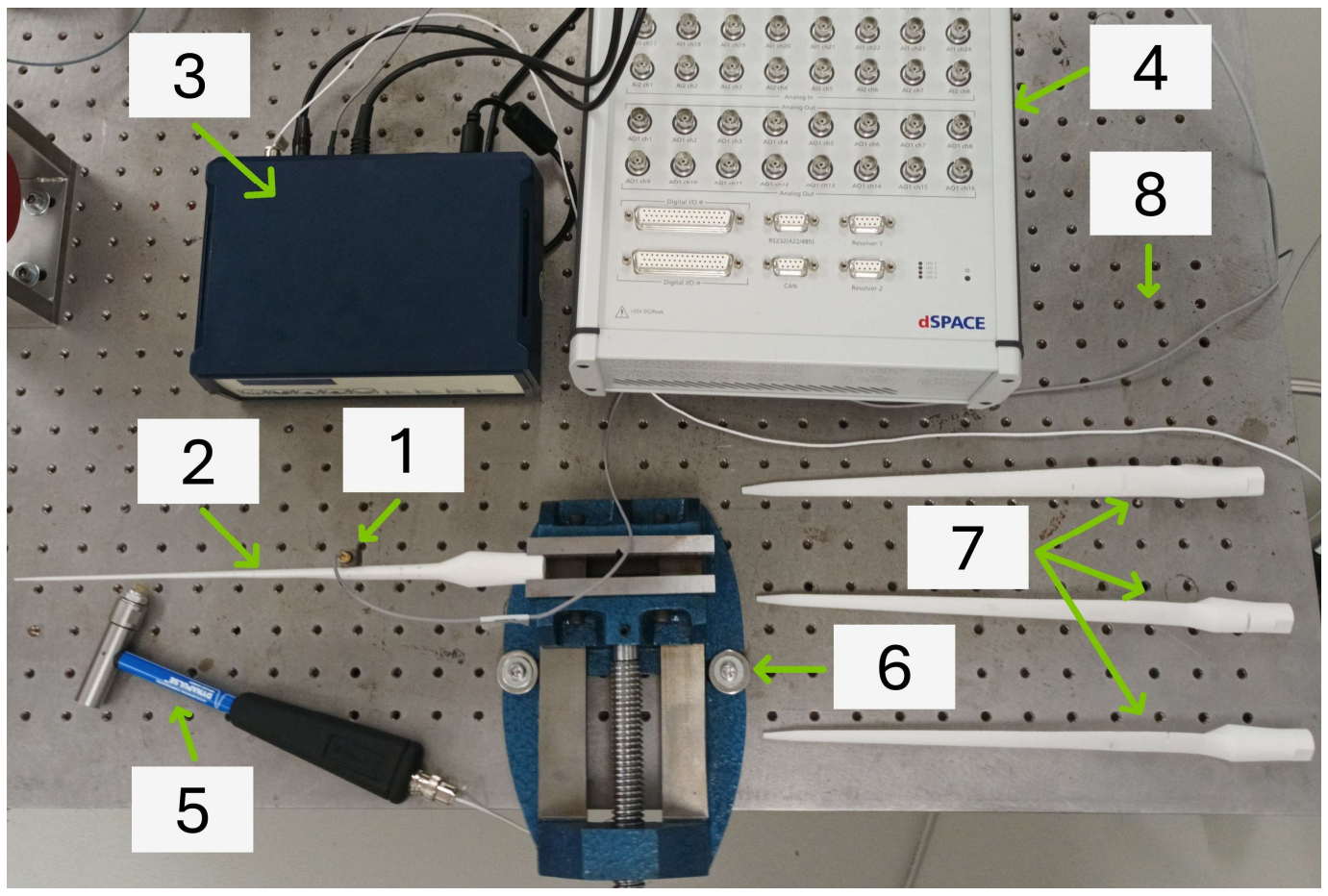

Fig. 2: Experimental setup: (1) Accelerometer 4507 B 004, (2) Tested blade, (3) Signal conditioning unit (PCB Piezotronics 482C Series), (4) Data acquisition system (dSpace MicroLabBox), (5) Dynapulse impulse hammer, (6) Bench vise, (7) Blade samples, (8) Anti-vibration table.

### 2.4 Data Preprocessing

Raw vibration data was truncated to 2 seconds post-impact and filtered using a fourth-order Butterworth low-pass filter with a cutoff frequency of 1000 Hz. The time and frequency domain features shown in Figure 6 were computed from the cropped and filtered signals. Moreover, the features were extracted using MATLAB's Diagnostic Feature Designer toolbox. Figure 6 shows the data density distribution for each feature across different damages. Finally, the features for training the models were chosen based on significant p-values ($p$ ă 0.05) from statistical tests, specifically ANOVA. These tests guarantee that the selected features exhibit strong discrimination between class labels, making them highly relevant for accurate classification and improving the overall model performance [12].

### 2.5 Data Processing

Extracted features were used to train the following classifiers: Random Forest (RF), Support Vector Machine (SVM), K-Nearest Neighbors (KNN), and Naive Bayes (NB) [10], [12]. Each Machine Learning (ML) model used normalized features, with data split into 70% training and 30% testing. The same training and testing set was used for each classifier, securing direct comparison between models.

## 3 RESULTS AND DISCUSSION

The modal analysis of the blade, performed through simulations, resulted in the displacement distributions along its length, shown in Figure 3b for the $x$-direction (bending modes) and Figure 3b for the $y$-direction (transverse modes). As observed for the first six modes, none exhibit pure bending or transverse behavior due to the blade's geometry. However, modes 1, 4, and 6 are predominantly bending, while mode 5 is primarily transverse, and modes 2 and 3 display transitional characteristics. These distinctions are crucial since the sensor used measures vibrations only along the $x$-axis. Additionally, Figure 3 highlights the sensor location and the corresponding displacements in each direction for its position. Although the sensor is suitably positioned for measuring modes 2 to 6, mode 1 exhibits minimal displacement at the sensor location. However, the decision was made to sacrifice its accurate capture, as the sensor's mass accounts for 13.6% of the blade's weight. To minimize the impact of added off-centered mass on the blade's dynamic behavior, the sensor was placed where its influence remained low while still allowing sufficient quantification of modes 1 to 6. Furthermore, it can be observed that damages 2 to 5 are located in positions where lateral displacements occur, whereas damage 1 is positioned near the base node of all vibration modes.

The analysis of the Frequency Response Functions (FRF) obtained from the hammer test simulations using the Finite Element Method in MATLAB revealed variations in the natural frequencies of vibration modes 3, 4, and 6 due to the introduced blade damages (Figure 4, dashed bars). However, modes 1, 2, and 5 exhibited minimal or no significant variation across different damages. As previously analyzed, mode 5 is predominantly transverse and is therefore not adequately captured by the sensor (both physical and simulated). Mode 1 exhibits insufficient displacement, while mode 2 is a transition mode closely related to mode 3, which, in this case, presents higher energy. A clear trend of frequency reduction was observed, in some cases reaching up to 4 Hz. This is consistent with the theoretical prediction that cracks reduce structural stiffness and lead to lower natural frequencies. An exception was Damage 5, which corresponds to a longitudinal defect on the leading edge, primarily affecting mass—2% reduction of healthy blade mass—rather than stiffness, leading to an increase in frequency because modal frequency is inversely related to mass; a decrease in mass typically leads to an increase in modal frequency. This is because the natural frequency of a system is proportional to the square root of the stiffness-to-mass ratio [4].

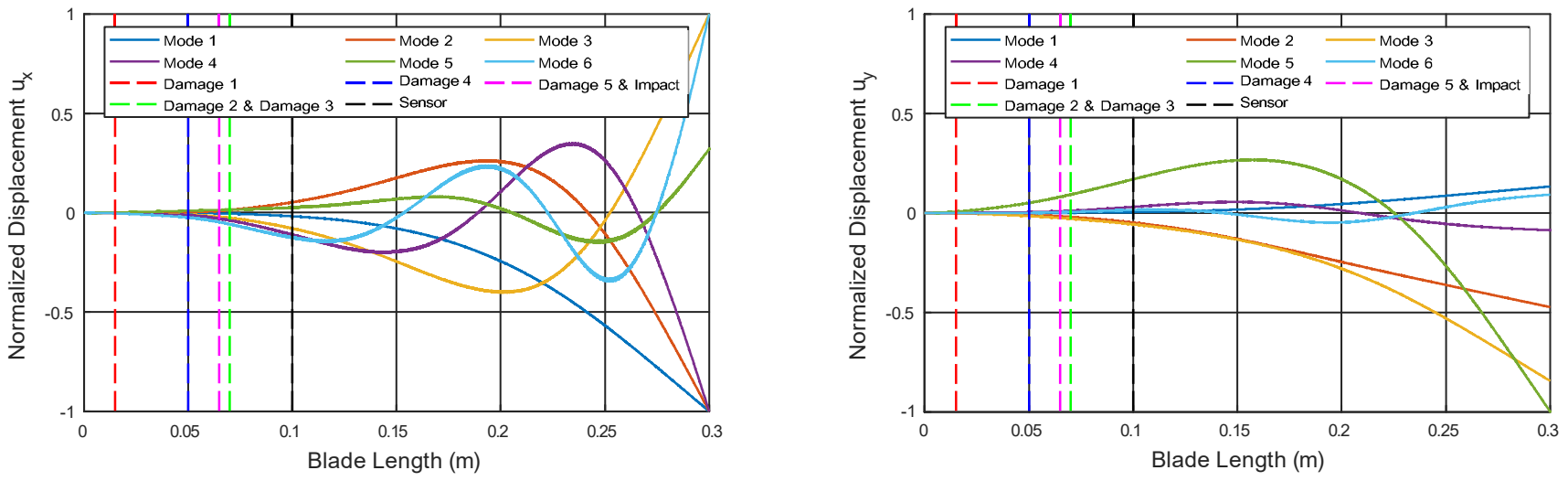

(a) (b)

Fig. 3: Normalized mode shapes highlight multiple damage scenarios (vertical dashed lines) and the sensor and impact location. (a) Variation of $u_x$ along the blade length. (b) Variation of $u_y$ along the blade length. Each vibration mode is normalized with respect to its corresponding counterpart on the orthogonal axis.

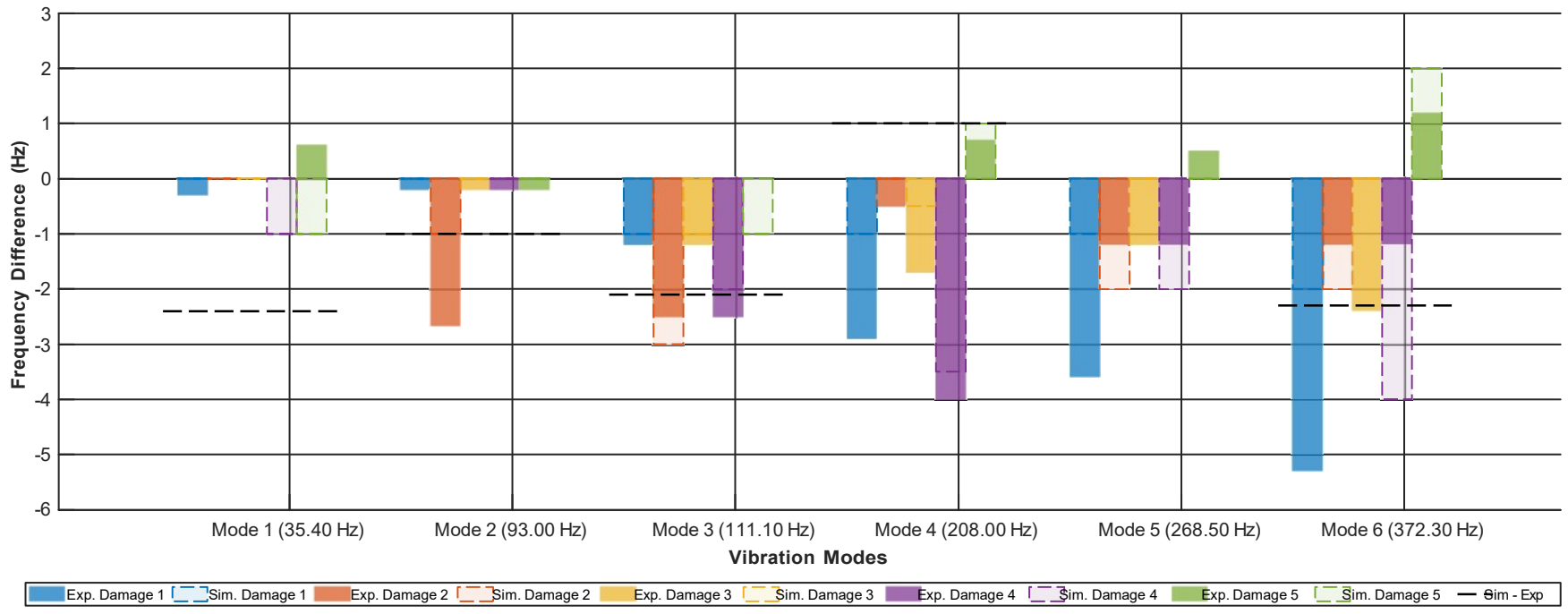

Fig. 4: Comparison of resonant frequency shifts for Modes 1–6: Damage blades vs. healthy blade. Experimental data (solid bars) and simulated data (dashed bars) are compared, with black dashed lines marking the discrepancy between the simulation and experiment for the healthy blade. Notably, the difference in Mode 5 exceeds 6 Hz.

The simulation results established a foundation for the manufacturing and testing of fault specimens. The FRF results for each of the 20 individual tests per damage sample (five samples per damage type) are presented in Figure 5. These results demonstrate consistency in resonance frequencies across samples for all tested damage types. The dashed blue lines represent the average frequencies of the vibration modes for each damage. For the healthy blade, the average frequencies for modes 1 to 6 were found to be [35.4, 93.0, 111.1, 207.5, 267.0, 372.3] Hz, respectively. Higher-order modes were not considered due to data inconsistency and noise affecting the quality of the FRF results.

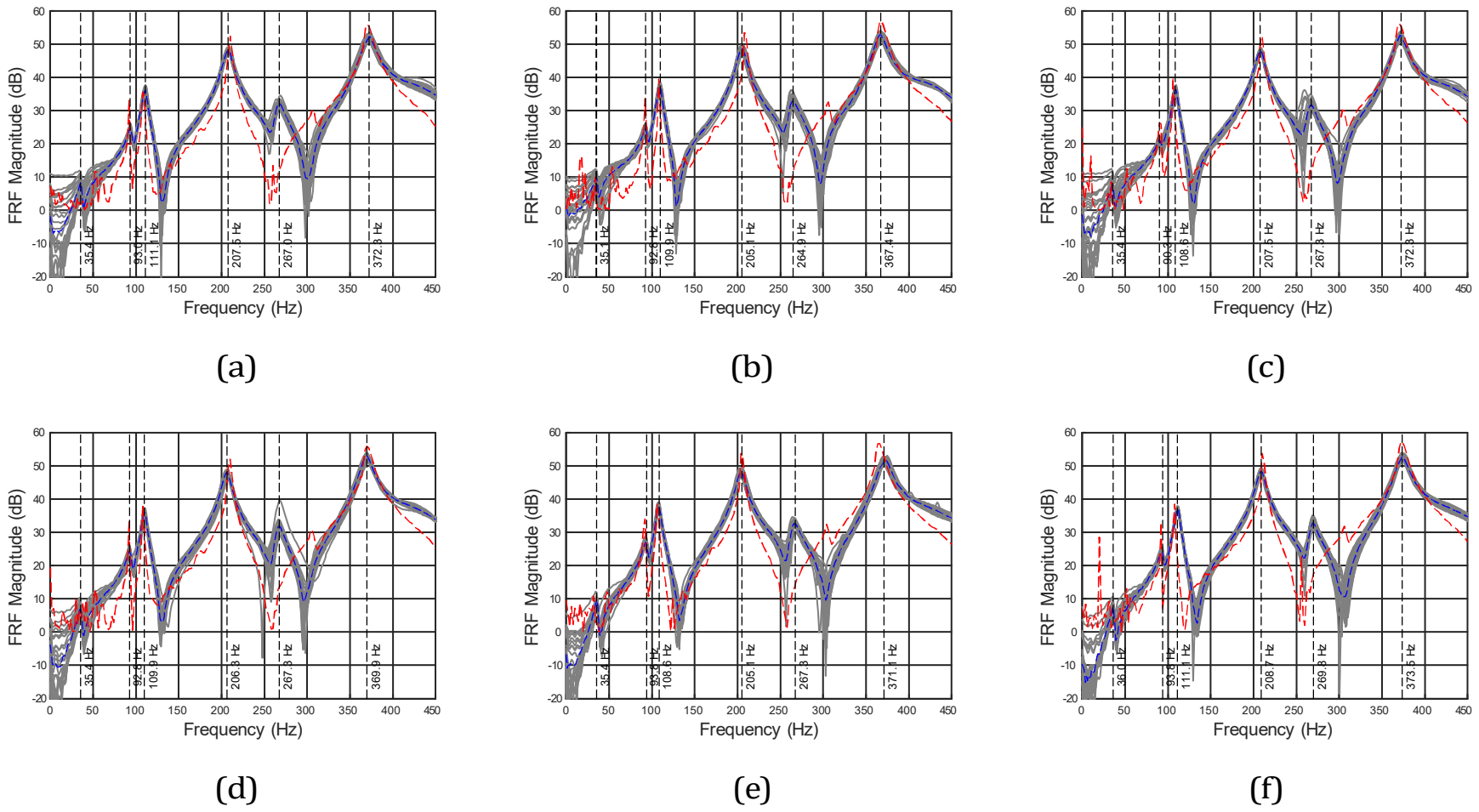

Fig. 5: Comparison of Frequency Response Functions (FRFs) for all individual experimental samples (gray), the experimental average (blue dashed), and simulated data (red dashed). Vertical black dashed lines indicate the average resonant frequencies in the experimental data. Subfigures correspond to: (a) Healthy, (b) Damage 1, (c) Damage 2, (d) Damage 3, (e) Damage 4, and (f) Damage 5.

Damages 1 to 5 exhibited variations in resonance frequencies compared to the healthy blade, consistent with simulation results (superimposed as dashed red lines in the subplots of Figure 5). Modes 2, 3, 4, and 6 correspond strongly with simulations. However, mode 5, which is predominantly a transverse mode, displayed a discrepancy in resonance frequency as in the simulated data. This deviation can be attributed to the use of a single-degree-of-freedom accelerometer, which does not fully capture this mode but still reflects traces of it in the signals due to the mode shapes influenced by the blade geometry

The absolute variations in resonance frequencies relative to the healthy blade (both experimental and simulated data) are presented in Figure 4, where a comparison with simulation results further confirms consistency. Modes 3, 4, and 6 exhibited the most significant variations, aligning with expectations since these modes are more affected by changes in structural stiffness due to damage. In Figure 4, black dashed lines indicate the differences between mode frequencies for simulated and experimental data. For mode 5, this difference is 37.5 Hz; however, it is not displayed in the figure due to space constraints. Nonetheless, a strong correspondence is observed for the other mode frequencies, with variations not exceeding 3.5 Hz between simulation and experiment.

Feature extraction in both time and frequency domains for training ML classifiers was performed using MATLAB's Diagnostic Feature Designer Toolbox.

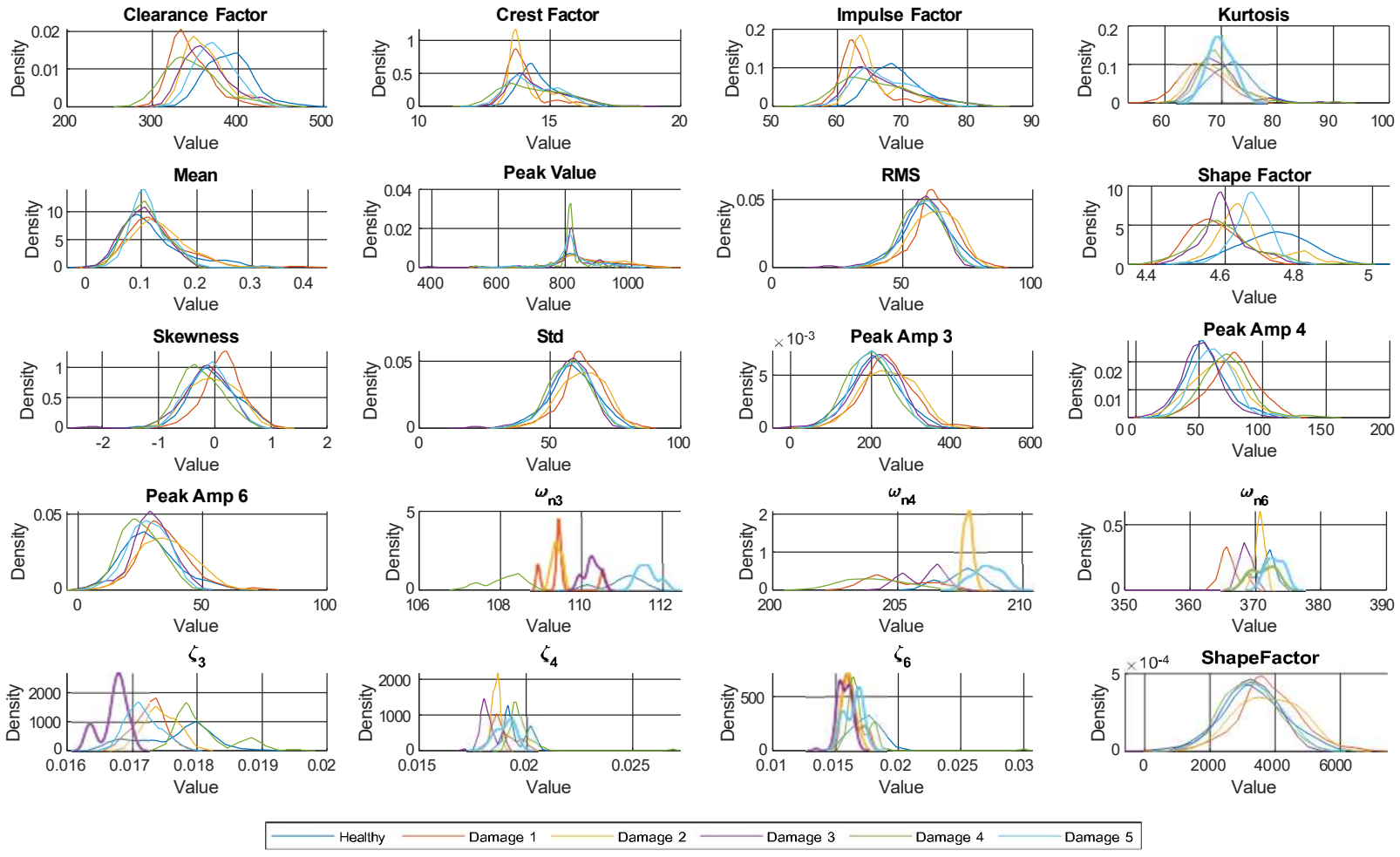

Fig. 6: Density distributions of feature values used for fault classification. The features $\omega_{n3}$, $\omega_{n4}$, $\omega_{n6}$, $\zeta_3$, $\zeta_4$, and *Shape Factor* exhibit the highest classification power based on the ANOVA test. Each colored line represents the density distribution of a specific feature across various damage types.

Figure 6 displays the density distributions of the selected features, showing clear separation trends in peak densities for some features. Ranking analysis using ANOVA tests determined the most discriminative features for fault classification, identifying $\omega_{n3}$, $\omega_{n4}$, $\omega_{n6}$, $\zeta_3$, $\zeta_4$, and *Shape Factor* as the most relevant in that order, where $\omega_n$ and $\zeta_n$ was extracted for the peaks of higher amplitude, specifically 111.1 Hz, 207.5 Hz, and 372.3 Hz, corresponding to modes 3, 4, and 6, respectively. This ranking aligns with the previous frequency analysis results, at least for the resonance frequencies of the indicated modes.

Four ML classifiers were trained and tested on the same dataset to evaluate their classification performance. All possible pairs of the two selected features were tested to determine the best feature combination. Table 2 highlights the top five feature combinations that achieved the highest classification accuracy. Among them, $\omega_{n3}$ and $\omega_{n4}$ consistently yielded the best results across all classifiers. The corresponding confusion matrices are presented in Figure 7.

Table 2: Accuracy (%) for the top 5 combinations of features in ML models.

| Features | SVM | KNN | NB | RF |
|---|---|---|---|---|
| $\omega_{n3}$ & $\omega_{n4}$ | 94.44 | 95.56 | 88.89 | 94.44 |
| $\omega_{n3}$ & $\omega_{n6}$ | 91.01 | 94.38 | 86.52 | 95.51 |
| $\omega_{n3}$ & $\zeta_3$ | 79.78 | 92.13 | 87.64 | 93.26 |
| $\omega_{n3}$ & $\zeta_4$ | 82.02 | 91.01 | 83.15 | 92.13 |
| $\omega_{n3}$ & *Shape Factor* | 87.64 | 93.26 | 73.03 | 92.13 |

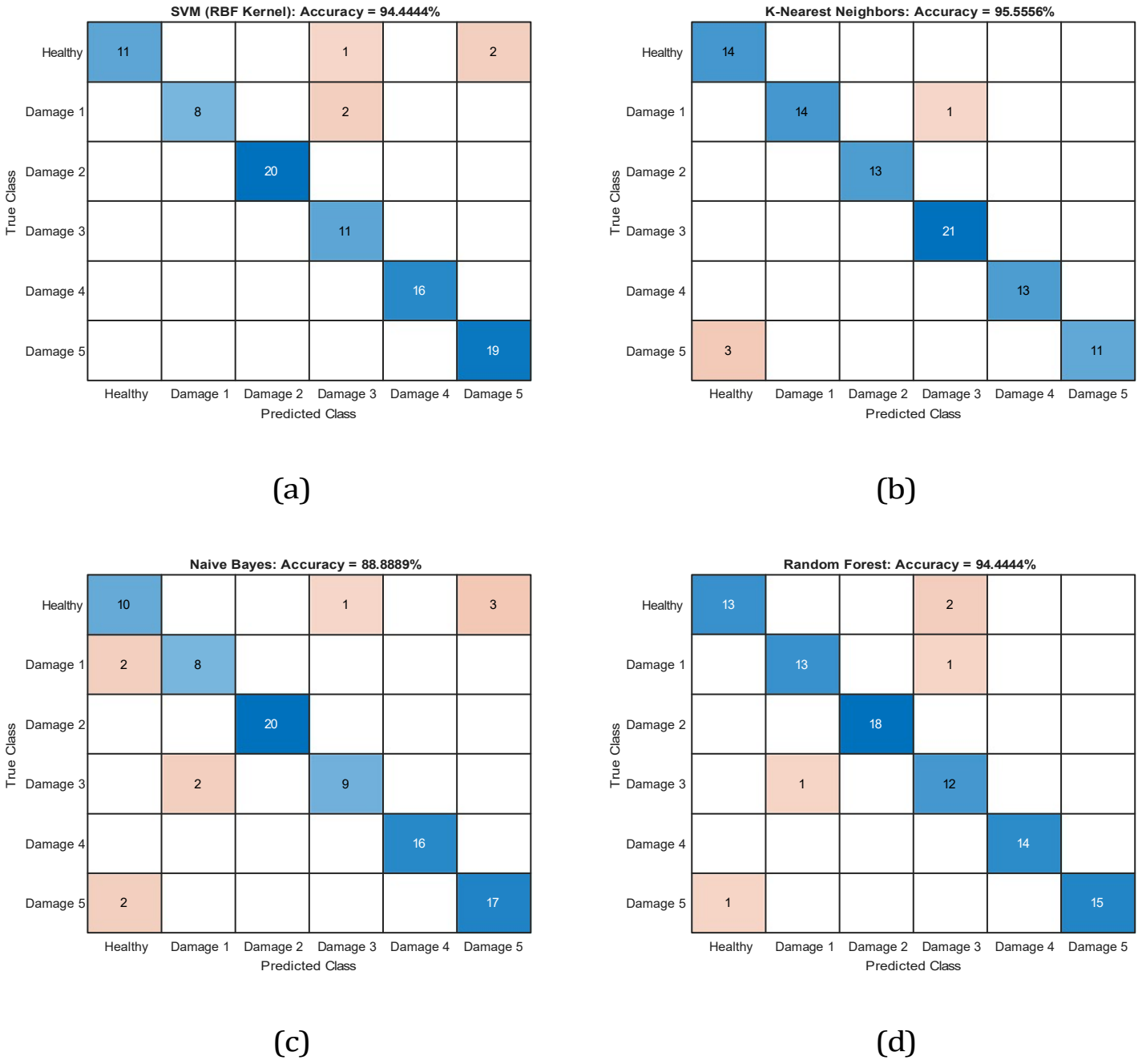

Fig. 7: Confusion matrices for the four ML classifiers trained on features $\omega_{n3}$ and $\omega_{n4}$, with the accuracy percentage of each model indicated. Each matrix displays the comparison between predicted and actual classes, highlighting the counts of true positives, true negatives, false positives, and false negatives.

# 4 CONCLUDING REMARKS AND FUTURE WORK

This study presents an integrated approach for fault detection in wind turbine blades using 3D-printed models, Finite Element Method simulations, and machine learning techniques. The results revealed that vibration modes 3, 4, and 6 are particularly sensitive to structural damage, with frequency shifts of up to 3 Hz observed compared to the healthy blades. Frequency-domain feature analysis identified $\omega_{n3}$, $\omega_{n4}$, and $\omega_{n6}$ as the most discriminative predictors, with machine learning classifiers, including SVM and KNN, achieving accuracies exceeding 94%. These findings demonstrate the effectiveness of the proposed methodology for damage detection.

The integration of FEM simulations with experimental data provides a reliable framework for fault diagnosis, validated by the experimental results in a controlled environment using 3D-printed models. This setup ensures high reproducibility and low implementation costs, making the methodology suitable for widespread adoption in the wind energy industry.

Future work will expand the experimental setup to include multi-blade configurations, enabling the study of cross-influences of damages between adjacent blades. Furthermore, detecting multiple faults within a single blade, such as cracks in different locations, will be explored to enhance the diagnostic capabil-ities. Sensor setups will be improved by incorporating multi-axis accelerometers to capture complex vibration modes better. Additionally, optimising data preprocessing techniques and integrating the methodology into real-time monitoring systems will be prioritized, enabling scalable, automated fault detection and contributing to predictive maintenance in wind turbines.

## References

1. Ali, W., El-Thalji, I., Giljarhus, K.E.T., Delimitis, A.: Classification analytics for wind turbine blade faults: Integrated signal analysis and machine learning approach. Energies **17**(23), 5856 (2024)
2. Brian, R.: Definition of a 5mw/61.5 m wind turbine blade reference model. Sandia National Laboratories: Albuquerque, NM, USA (2013)
3. Garc´ıa Marquez, F.P., Peco Chac´on, A.M.: A review of non-destructive testing on wind turbines blades (2020). https://doi.org/10.1016/j.renene.2020.07.145
4. Jayswal, S., Bhattu, A.: Structural and modal analysis of small wind turbine blade using three different materials. In: Materials Today: Proceedings. vol. 72 (2023). https://doi.org/10.1016/j.matpr.2022.09.329
5. Kaewniam, P., Cao, M., Alkayem, N.F., Li, D., Manoach, E.: Recent advances in damage detection of wind turbine blades: A state-of-the-art review (2022). https://doi.org/10.1016/j.rser.2022.112723
6. Li, D., Ho, S.C.M., Song, G., Ren, L., Li, H.: A review of damage detection methods for wind turbine blades (2015). https://doi.org/10.1088/0964-1726/24/3/033001
7. Lorenzo, E.D., Petrone, G., Manzato, S., Peeters, B., Desmet, W., Marulo, F.: Damage detection in wind turbine blades by using operational modal analysis. Structural Health Monitoring **15**(3) (2016). https://doi.org/10.1177/1475921716642748
8. Mishnaevsky, L.: Root Causes and Mechanisms of Failure of Wind Turbine Blades: Overview (2022). https://doi.org/10.3390/ma15092959
9. Morettini, G., Palmieri, M., Capponi, L., Landi, L.: Comprehensive characterization of mechanical and physical properties of PLA structures printed by FFF-3D-printing process in different directions. Progress in Additive Manufacturing **7**(5) (2022). https://doi.org/10.1007/s40964-022-00285-8
10. Ogaili, A.A.F., Jaber, A.A., Hamzah, M.N.: A methodological approach for detecting multiple faults in wind turbine blades based on vibration signals and machine learning. Curved and Layered Structures **10**(1) (2023). https://doi.org/10.1515/cls-2022-0214
11. Öteyaka, M.Ö., Çakir, F.H., Sofuo˘glu, M.A.: Effect of infill pattern and ratio on the flexural and vibration damping characteristics of FDM printed PLA specimens. Materials Today Communications **33** (2022). https://doi.org/10.1016/j.mtcomm.2022.104912
12. Rangel-Rodriguez, A.H., Granados-Lieberman, D., Amezquita-Sanchez, J.P., Bueno-Lopez, M., Valtierra-Rodriguez, M.: Analysis of Vibration Signals Based on Machine Learning for Crack Detection in a Low-Power Wind Turbine. Entropy **25**(8) (2023). https://doi.org/10.3390/e25081188
13. Wang, M.H., Lu, S.D., Hsieh, C.C., Hung, C.C.: Fault Detection of Wind Turbine Blades Using Multi-Channel CNN. Sustainability (Switzerland) **14**(3) (2022). https://doi.org/10.3390/su14031781